%% file: ssrr_2018_aerial_arxiv_preprint.tex
\documentclass[letterpaper,10pt,conference]{ieeeconf}
\IEEEoverridecommandlockouts

\usepackage[numbers,sort&compress]{natbib}
\usepackage{graphicx}
\usepackage{epsfig}
\usepackage{verbatim}
\usepackage{color}
\usepackage{soul}
\usepackage[font=footnotesize]{caption}
\usepackage{amssymb}
\usepackage{amsmath,bm}
\usepackage[english]{babel}
\usepackage[T1]{fontenc}
\usepackage[utf8]{inputenc}
\usepackage{xspace}
\usepackage{todonotes}
\usepackage{epstopdf}
\usepackage{hyperref}
\usepackage{url}
\usepackage[nolist,nohyperlinks]{acronym}
\usepackage{tikz}
\usepackage{pgfplots}
\usepackage{subcaption}
\usepackage{makecell}

\title{Aerial-Ground collaborative sensing: Third-Person view for teleoperation}

\author{Abel Gawel, Yukai Lin, Th\'eodore Koutros, Roland Siegwart and Cesar Cadena
\thanks{*Authors are with the Autonomous Systems Lab, ETH Zurich, {\tt\small \{gawela, rsiegwart, cesarc\}@ethz.ch, \tt\small \{linyuk, koutrost\}@student.ethz.ch}}%
}

\begin{document}

\onecolumn
{\large
\input{preface.tex}
\par}
\normalsize
\twocolumn
\clearpage
\setcounter{page}{1}
\maketitle

\begin{abstract}
Rapid deployment and operation are key requirements in time critical application, such as \ac{SaR}.
Efficiently teleoperated ground robots can support first-responders in such situations.
However, first-person view teleoperation is sub-optimal in difficult terrains, while a third-person perspective can drastically increase teleoperation performance.
Here, we propose a \ac{MAV}-based system that can autonomously provide third-person perspective to ground robots. 
While our approach is based on local visual servoing, it further leverages the global localization of several ground robots to seamlessly transfer between these ground robots in GPS-denied environments.
Therewith one \ac{MAV} can support multiple ground robots on a demand basis.
Furthermore, our system enables different visual detection regimes, and enhanced operability, and return-home functionality.
We evaluate our system in real-world \ac{SaR} scenarios.
\end{abstract}

\IEEEpeerreviewmaketitle

\section{Introduction}
Effective teleoperation is a key requirement for many contemporary \ac{UGV} systems. 
%
%
Usually, these systems are teleoperated in a first-person perspective, using on-board cameras and further sensors of the robots.
While this is sufficient in easy terrain, the teleoperation task can become cumbersome in challenging terrains, where narrow passages or obstacles need to be traversed.
%
%
In these situation not only the overall progress of a mission can be compromised but also the integrity of the vehicle in use. 

\ac{MAV}s on the other hand offer rapid speeds and a higher point of view, giving them superior performance as flying cameras.
However, often it is not sufficient to use \ac{MAV}s alone in such scenarios as their operation times and payload are typically more constrained than of \ac{UGV}s.
Here, \ac{UGV}s can complement the \ac{MAV} capabilities with manipulators, and further sensors for close interaction with the environment.
%
%
%
%
%

\begin{figure}
\centering
\includegraphics[width = 0.9\columnwidth]{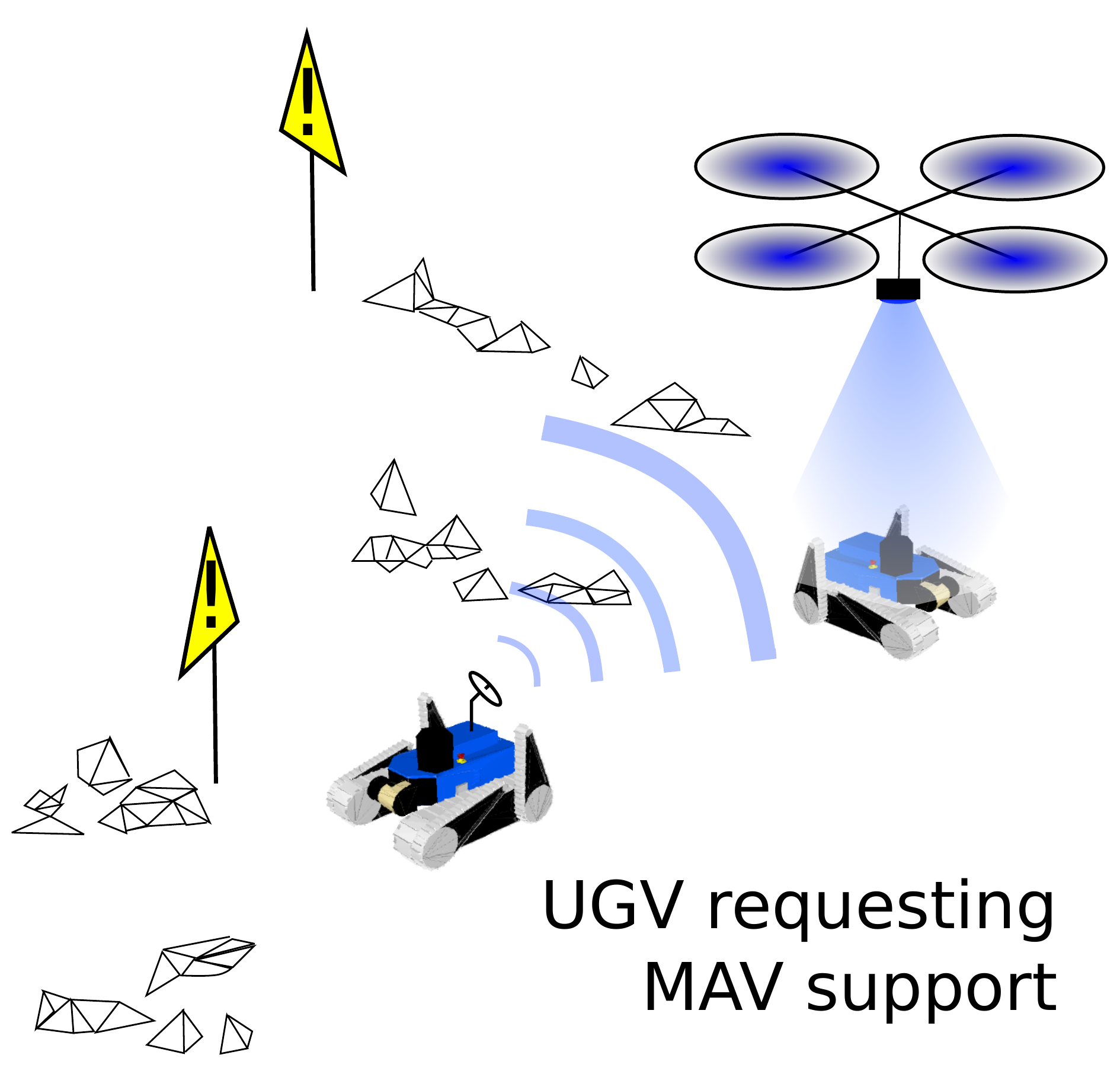}
\caption{Illustration of our proposed system. \ac{MAV} supporting multiple UGVs in teleoperating difficult terrain by providing third-person view on request.}
\label{fig:pictogram}
\vspace{-1.0em}
\end{figure}

Therefore, the combination of the individual robots' strengths in an integrated system can be a fruitful avenue~\cite{saakes2013teleoperating, gawel20173d, gawel2016structure, gawel2017aerial}.
Recent works have shown that a third-person perspective can prove efficient to support robot operators in such scenarios~\cite{saakes2013teleoperating, cantelli2013uav}, e.g., by using \ac{MAV}s as flying cameras.
Unfortunately, these systems are limited to off-board computations, structured and well illuminated scenarios, single-\ac{UGV} support, tag-based detection \cite{olson2011tags}, and local-following only.
Here, we extend our prior work~\cite{gawel2017aerial} and propose an automatic system based on a \ac{MAV} that can overcome these limitations and support multiple \ac{UGV}s in difficult teleoperation tasks.
In our system, the \ac{MAV} serves as a flying camera that can autonomously follow a \ac{UGV} and provide a third-person view, as illustrated in Fig.~\ref{fig:pictogram}.
Our system is designed to operate in GPS-denied environments, and therefore we assume no external localization system.
Therefore, the \ac{MAV} localization is loosely coupled with the \ac{UGV} localization, and can therewith support multiple \ac{UGV}s.
The \ac{MAV} interfaces with the LiDAR-based global localization of the \ac{UGV}s allowing for traveling among them on a demand basis.
%
%
Visual-inertial sensing onboard the \ac{MAV} allows it to navigate between multiple \ac{UGV}s and assist teleoperation.
Our system operates fully autonomous, freeing operators from the need to separately control the flying camera.
The system has been tested in real-word experiments in challenging indoor \ac{SaR} scenarios.
To the best of our knowledge, this is the first system to integrate all these functionalities.

This paper presents the following contributions:
\begin{itemize}
\item A system for relative localization, visual servoing, and control of MAV for third-person view teleoperation.
\item Feature-based detection of different \ac{UGV} types.
\item An integration with \ac{UGV} localization that enables global localization and transfer between different \ac{UGV}s in GPS-denied environments.
\item Experimental evaluation in challenging real-world \ac{SaR} scenarios.
\end{itemize}
This paper is organized as follows;
In Section~\ref{sec:rel_work}, we review the state of the art on \ac{MAV}-\ac{UGV} collaboration, visual servoing, and third-person view teleoperation.
Section~\ref{sec:method} describes our integrated system, and its evaluation is presented in Section~\ref{sec:experiments}.
Finally, we conclude our findings in Section~\ref{sec:conclusion}.
\section{Related work}
\label{sec:rel_work}
%
The topic of \ac{MAV}-\ac{UGV} cooperation has received increasing attention in the last decade with the advent of affordable \ac{MAV}s and increased onboard processing power~\cite{rudol2008micro, cantelli2013uav, saakes2013teleoperating, gawel20173d, gawel2016structure}.
Many contemporary works focus on enhancing the \ac{UGV}'s perception using \ac{MAV}s~\cite{cantelli2013uav, cantelli2017uav, saska2012cooperative, saakes2013teleoperating, claret2016teleoperating, gawel20173d, gawel2016structure, hui2013autonomous, michael2014collaborative}.
A related functionality is to use \ac{MAV}s for interaction, e.g., object picking and transportation~\cite{bernard2011autonomous, gawel2017aerial, bahnemann2017decentralized}.
Typically, the \ac{UGV} and the \ac{MAV} need to be co-localized and controlled in a shared reference frame~\cite{chaumette2014visual}.
This is commonly solved using visual servoing, i.e., visual detection between the robots and applying a suitable control regime~\cite{lee2012autonomous}.

The detection in visual servoing systems, is often based on special markers~\cite{cantelli2013uav, cantelli2017uav, rudol2008micro, saska2012cooperative, saakes2013teleoperating}, simple visual color blob detection~\cite{gawel2017aerial, bahnemann2017decentralized, hui2013autonomous}, or visual feature-based detection~\cite{pestana2014computer}.
Special markers, e.g., Apriltags~\cite{olson2011tags}, are a well established technique for robust visual-marker detection, and have superseded LED-based detection in recent years~\cite{cantelli2017uav}.
In more general visual servoing cases, the \ac{UGV} cannot be augmented with specific markers.
Hence, visual detection is still needed in such cases.

Another option is to localize the robots in a common map without direct detection~\cite{michael2014collaborative, gawel20173d, gawel2016structure}.
Here, both robots are equipped with additional LiDAR, or camera sensors for precise localization and collaborative mapping.
However, \citet{michael2014collaborative} state that additional direct detection between the robots would be desirable for direct collaboration.

The control for visual servoing is mainly on trajectory or waypoint tracking.
In recent years, the control problem for \ac{MAV}s has been intensively investigated.
A classic PID controller performs better than an LQ controller due to the model imperfections ~\cite{Bouabdallah2004pid}.
A nonlinear tracking controller often used is shown to have desirable closed loop properties with global stability~\cite{Lee2010geometric}. 
To exert state and input constraints, \ac{MPC} has been introduced and employed to control the \ac{MAV}, and it was further shown that Nonlinear \ac{MPC} performs better than linear \ac{MPC} in disturbance rejection and tracking performance~\cite{Kamel2017linear}.
%

Multiple works have shown that a third-person view can prove useful in teleoperation tasks~\cite{saakes2013teleoperating, burigat2017mobile, minaeian2016vision}.
While~\cite{burigat2017mobile} use \emph{3D} maps as they could be build from on-board sensing of the \ac{UGV}, \cite{minaeian2016vision} show the use of a \ac{MAV} as an external camera for assisted manipulation.
However, \cite{minaeian2016vision} focus on a static target, and \cite{burigat2017mobile} require additional \emph{3D} sensing on the \ac{UGV} to render a third-person view from the local sensing.
The most similar work to ours is~\cite{saakes2013teleoperating}.
Here, the authors focus on evaluating the assisted teleoperation by a \ac{MAV}-based system in \ac{SaR} scenarios.
While the system integration and development has been simplified and limited in order to perform the user-studies.
Some of those are off-board computing, single-\ac{UGV} support, and simplified environmental conditions.

Cooperation between multiple \ac{UGV}s and \ac{UAV}s has been researched in the last decades. 
An early work of \cite{Sukhatme2001experiments} studies the cooperation of two \ac{UGV}s and one aerial robot regarding localization of the aerial robot by visually communicating and locating with the ground robot.
\citet{Hsieh2007adaptive} demonstrated multi-agent tasking and provided cooperative control
strategies for search, identification, and localization of targets.
The survey of \cite{Waslander2013unmanned} identifies several open problems on \ac{UGV}/\ac{UAV} cooperation including vehicle autonomy, and integrated control.

This paper focuses on lifting simplifications of the current state of the art by extending our previous work on \ac{MAV} object detection and picking~\cite{gawel2017aerial, bahnemann2017decentralized}, adding dedicated multi-\ac{UGV} support, and visual-feature based object detection for \ac{MAV}-assisted teleoperation.
We use state of the art Nonlinear \ac{MPC} on the \ac{MAV}.
Furthermore, we demonstrate the system in perceptually difficult indoor \ac{SaR} scenarios and perform computations on-board the \ac{MAV} which ensures safe operation in case of network failures.

\section{Third-person View Teleoperation}
\label{sec:method}
\begin{figure}
\centering
\includegraphics[width = 0.9\columnwidth]{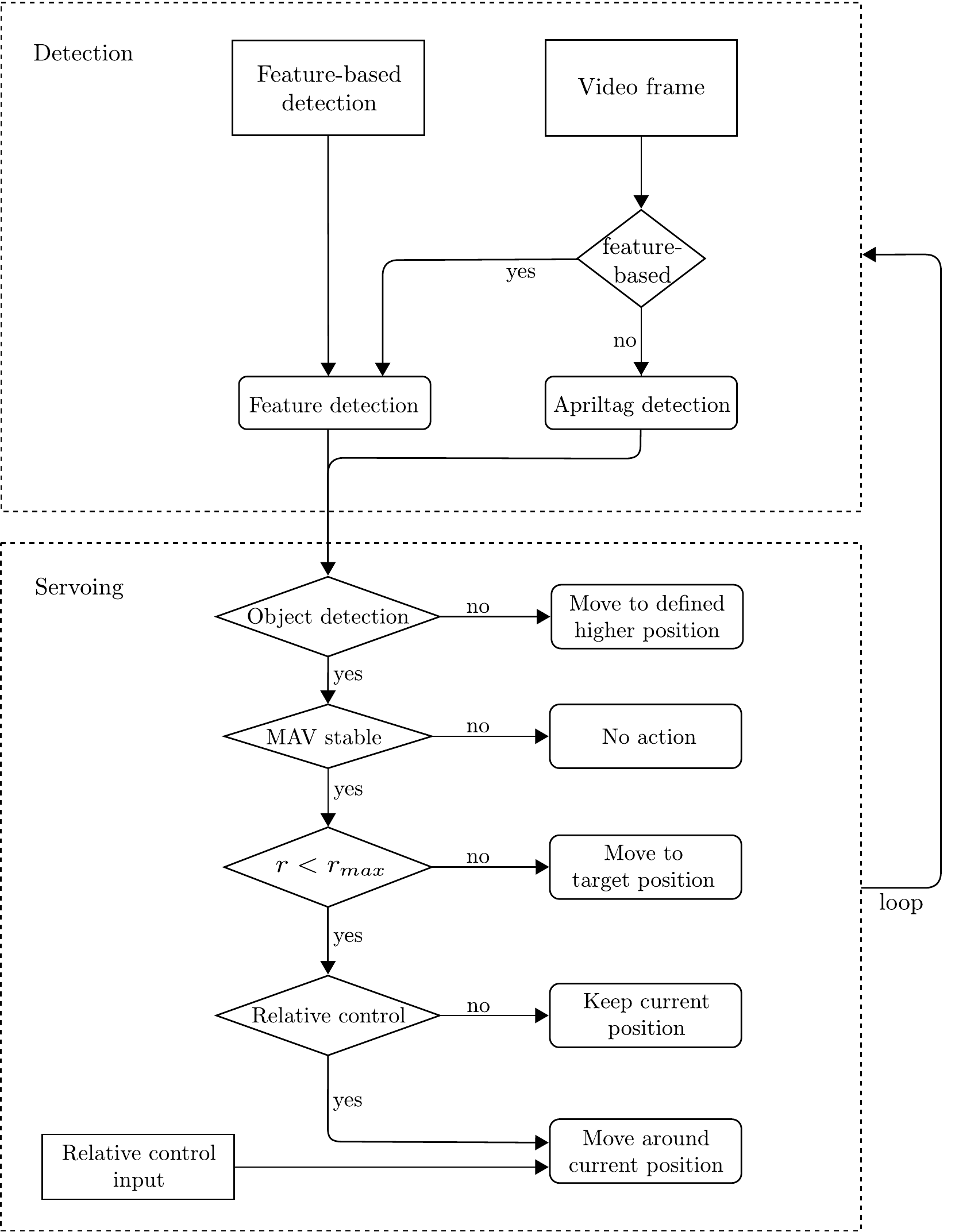}
\caption{\ac{MAV} detection and servoing strategy. The upper shows the object detection using either the Apriltag detection or visual feature-based detection given object image. The lower block receives the detection results and performs the servoing strategy accordingly.}
\label{fig:servoing_flowchart}
\vspace{-1.0em}
\end{figure}

Here, we present our \ac{MAV} assisted third-person view system for \ac{UGV} teleoperation.
The system is based on object detection, visual servoing, \ac{VIO}, and interfacing with the \ac{UGV}s' global localization to operate in GPS-denied environments.
A system overview is depicted in Fig.~\ref{fig:servoing_flowchart}.
As illustrated in the upper part of Fig.~\ref{fig:servoing_flowchart}, the detector uses either tag-based detection~\cite{olson2011tags},
or visual feature-based object detection, based on an initially captured image of the \ac{UGV}~\citep{leutenegger2011brisk}.%
The detections are then sent to the visual servoing, which computes the \ac{MAV}'s relative, and target poses, and applies the necessary control to the \ac{MAV} for hovering above the \ac{UGV}.
Finally, upon successful relative localization of the \ac{MAV} with respect to the \ac{UGV}, the global frames of \ac{UGV} and \ac{VIO} are aligned for global localization of the \ac{MAV}, and enabling transfer among multiple \ac{UGV}s.
\begin{figure}
\centering
\includegraphics[width = 0.9\columnwidth]{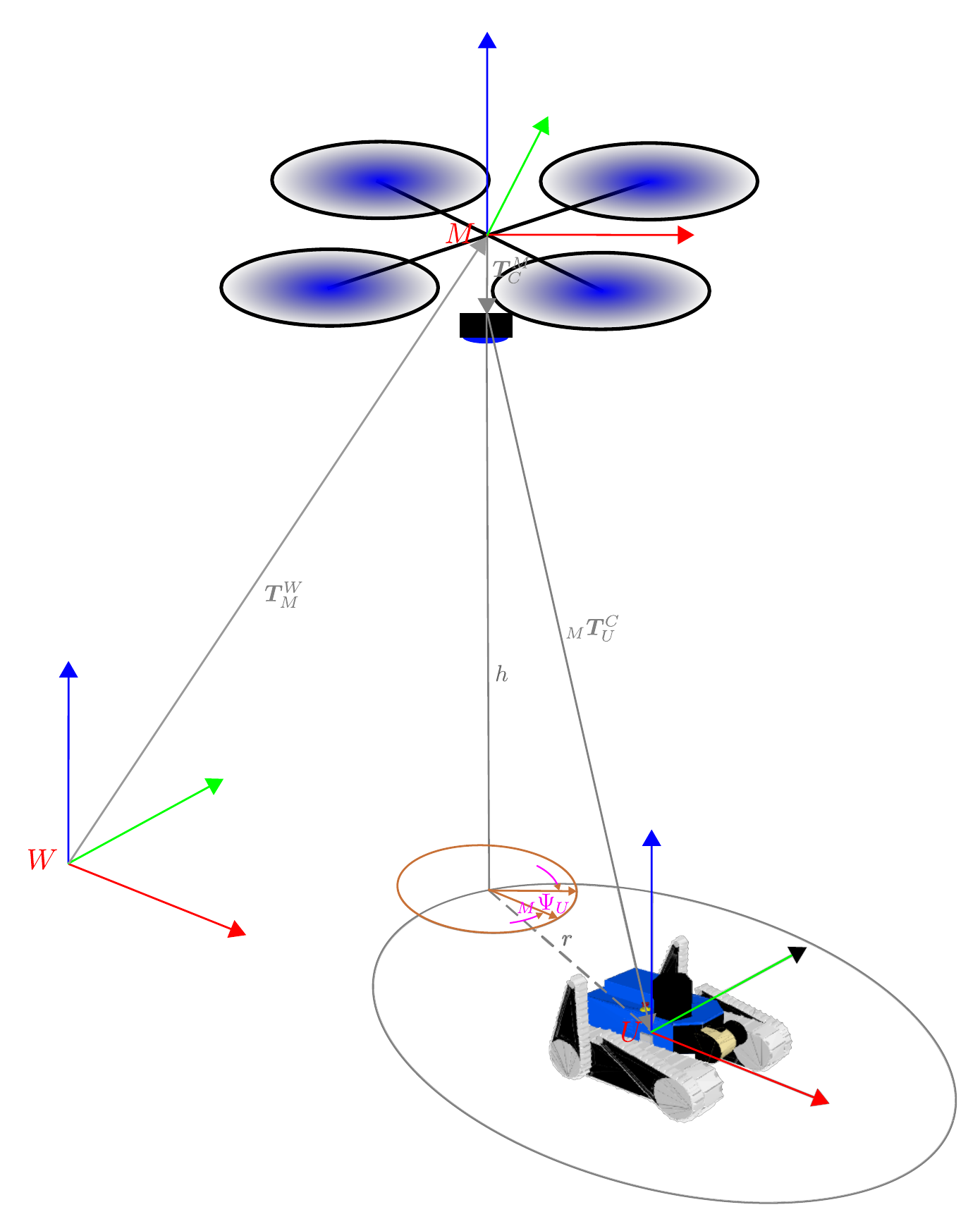}
\caption{Schematic overview of used frames, and transformation chain between \ac{MAV}, and \ac{UGV}. For visualization purposes, the coordinate system on the camera $C$ was omitted.}
\label{fig:drone_frame}
\vspace{-1.0em}
\end{figure}

\subsection{Visual Servoing}
The servoing strategy is agnostic to the used detection regime.
Either the pose of the detected tag, or the pose of the smallest quadrilateral box containing the target object when using feature-based \ac{UGV} detection are forwarded to the visual servoing algorithm.
We represent transformations $\bm{T}$ in \emph{SE3}, consisting of position $\bm{p}$, and orientation in roll $\phi$, pitch $\theta$, and, yaw $\Psi$ in quaternion form.
The algorithm first estimates the relative pose between \ac{MAV} and \ac{UGV} $_{M}\bm{T}_{U}^C$ in the camera frame $C$.
%
%
%
%

%
Here, we only consider rotations in yaw.
Rotations in pitch and roll are assumed negligible, since the \ac{MAV} can only remain stable when it is stationary hovering over the \ac{UGV}, and takes actions only in static hovering mode.
%

Besides relative localization, the \ac{MAV} also localizes itself in the world using \ac{VIO}, as implemented in ~\cite{bloesch2015robust}.
The position controller receives commands in world coordinates.
We therefore need to apply a transformation chain to the relative localization.
The \ac{MAV}'s pose $\bm{T}_{M}^W$ is represented in the world frame $W$.
Thus, ${_{M}}\bm{T}_{U}^C$ is then transformed into the world frame, using the \ac{VIO} estimate $\bm{T}_{M}^W$, and the extrinsic calibration between \ac{MAV} IMU and camera $\bm{T}_{C}^M$, forming the relative transformation $_{M}\bm{T}_{U}^W$ between \ac{MAV} and \ac{UGV} in world coordinates, i.e., 
\begin{equation}
{_{M}}\bm{T}_{U}^W = \bm{T}_{M}^W \bm{T}_{C}^M {_{M}}\bm{T}_{U}^C
\end{equation}
This is also illustrated in Fig.~\ref{fig:drone_frame}.

\begin{figure}
\centering
\includegraphics[width = \columnwidth]{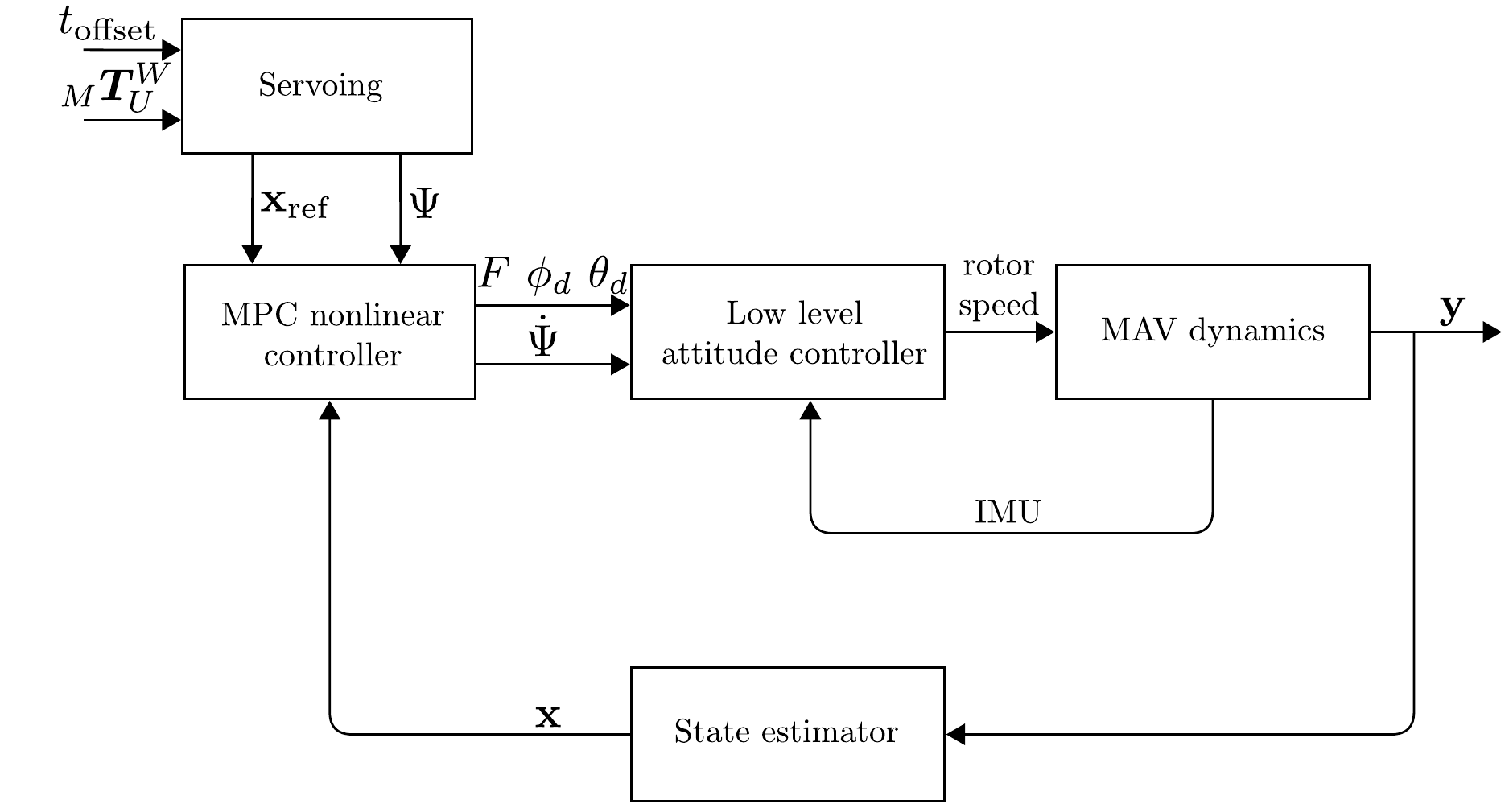}
\caption{Cascade controller structure for multi-rotor system. Here, the measured \ac{MAV} state is denoted by $\bm{\mathbf{y}}$.}
\label{fig:controller}
\vspace{-1.0em}
\end{figure}

The localization result is then sent to an \ac{MPC}~\cite{kamel2017model} for position control.
Here, the \ac{MPC} acts to keep the \ac{MAV} within a circle of radius $r < r_{max}$ in a height $h$ above the center of the \ac{UGV}, and identical yaw orientation, i.e., ${_M\Psi_U}=0$.
Furthermore, a user-controllable offset translation $\bm{t}_{\text{offset}}$ in \ac{UGV} body coordinates can be added.
%
The controller structure is shown in Fig.~\ref{fig:controller}.
The position command is handled by the \ac{MPC} nonlinear controller to generate desired motion control. 
The optimization problem for the \ac{MPC} is formulated as
\begin{equation}
\begin{split}
\min\limits_{\bm{\mathbf{u}},\bm{\mathbf{x}}}&\int_{t=t_0}^\tau
(\bm{\mathbf{x}}(t)-\bm{\mathbf{x}}_{\text{ref}}(t))^T \bm{Q}_\mathbf{x} (\bm{\mathbf{x}}(t)-\bm{\mathbf{x}}_{\text{ref}}(t))\\
&+
(\bm{\mathbf{u}}(t)-\bm{\mathbf{u}}_{\text{ref}}(t))^T \bm{R}_\mathbf{u} (\bm{\mathbf{u}}(t)-\bm{\mathbf{u}}_{\text{ref}}(t))dt\\
&+
(\bm{\mathbf{x}}(T)-\bm{\mathbf{x}}_{\text{ref}}(T))^T \bm{P} (\bm{\mathbf{x}}(T)-\bm{\mathbf{x}}_{\text{ref}}(T))\\
&\text{subject to } 
\begin{split}
&\dot{\bm{\mathbf{x}}} = \bm{f}(\bm{\mathbf{x}},\bm{\mathbf{u}});\\
&\bm{\mathbf{u}}(t)\in \mathbb{U};\\
&\bm{\mathbf{x}}(0)=\bm{\mathbf{x}}(t_0)
\end{split}
\end{split}
\end{equation}
in the time interval $t\in[t_0,\tau]$.
Here, $\bm{Q}_\mathbf{x}\geq 0$ is the penalty on the state error, $\bm{R}_\mathbf{u}>0$ is the penalty on control input error and P is the terminal state error.
The state vector $\bm{\mathbf{x}} = [\bm{p}^T, \bm{v}^T,\phi, \theta]^T$ represents the position, velocity, roll and pitch angle of the \ac{MAV} and input vector $\bm{\mathbf{u}} = [\phi_d,\theta_d, F]^T$ consists of control input of roll, pitch angle and thrust force $F$.
The desired state and steady state input are denoted as
$\bm{\mathbf{x}}_{\text{ref}}$ and $\bm{\mathbf{u}}_{\text{ref}}$.
Finally, the result is sent to a low level attitude controller to generate desired rotor speed control.

To yield more robust system performance, the control action is sent only when the \ac{MAV} is stable, i.e. hovering or moving constantly.
The advantage of using \ac{VIO} are stable position control even without visual servoing, and safety (and recovery) measures in the case of visual tracking failures.
%

%
\subsection{Multi-UGV support}
In our system, the \ac{UGV}s are performing LiDAR-based \emph{3D} SLAM~\cite{dube20163d}, enabling them to globally localize in a common frame.
The SLAM system operates on a pose-graph basis, registering ICP LiDAR scan alignments, and odometry.
Furthermore, open-loop drift is compensated upon loop-closures using ICP and a low drift assumption.

Upon start, we initialize all robots, including \ac{MAV} in the origin of the world frame, yielding a shared reference frame.
This common frame is maintained via LiDAR-based SLAM for the \ac{UGV}s, i.e., estimating their pose $\bm{T}_{U}^W$ and \ac{VIO} for the \ac{MAV} $\bm{T}_{M}^W$.
The knowledge of the poses of \ac{UGV}s in the world frame allow us to calculate the relative transform $_{U,i}\bm{T}_{U,j}$ between two \ac{UGV}s $i$ and $j$.
This enables robot-to-robot third-person-view transfers.
Passed to the controller of the \ac{MAV}, the \ac{MAV} can transfer between multiple \ac{UGV}s, before switching back to visual servoing.

One important characteristic of this hybrid system is different drift characteristics in the LiDAR-based localization and the \ac{VIO} for \ac{UGV} and \ac{MAV}, respectively.
In our system, we give higher confidence in the accuracy of LiDAR-based localization and therefore reset $\bm{T}_{M}^W$ on transition events towards the \ac{UGV} estimate, i.e.,
\begin{equation}
\bm{T}_{M}^W := \bm{T}_{U}^W ({_{M}}\bm{T}_{U}^W)^{-1}
\end{equation}
The relative drift between \ac{VIO} and LiDAR-based SLAM are then compensated through the visual servoing.

The return-home functionality is realised with the same detection regime.
When the \ac{MAV} concludes its mission, the operator can request it to return from visual servoing above a \ac{UGV} to a home position in the world frame, e.g., the starting point.
The \ac{MAV} will thus return to the home position using \ac{VIO} and descend to the ground.
If high accuracy is required upon landing, the home position can be equipped with a visual tag, such that the visual servoing will correct for odometry drift before landing.
\section{Experiments}
\label{sec:experiments}
We evaluate our system in two different indoor experiments and record its performance.
Our evaluation focuses on the visual servoing performance and the proposed multi-\ac{UGV} interface and transfer with the \ac{MAV}.
We furthermore demonstrate the effectiveness of the third-person teleoperation in a realistic industrial \ac{SaR} scenario.
The benefits of third-person view teleoperation has been extensively evaluated and concluded in \cite{saakes2013teleoperating, burigat2017mobile, minaeian2016vision}, and further evaluations are therefore outside the scope of this paper.
\subsection{Experimental setup}
\begin{figure*}[!htb]
\centering
\begin{subfigure}[c]{0.45\textwidth}
{\includegraphics[height=5cm]{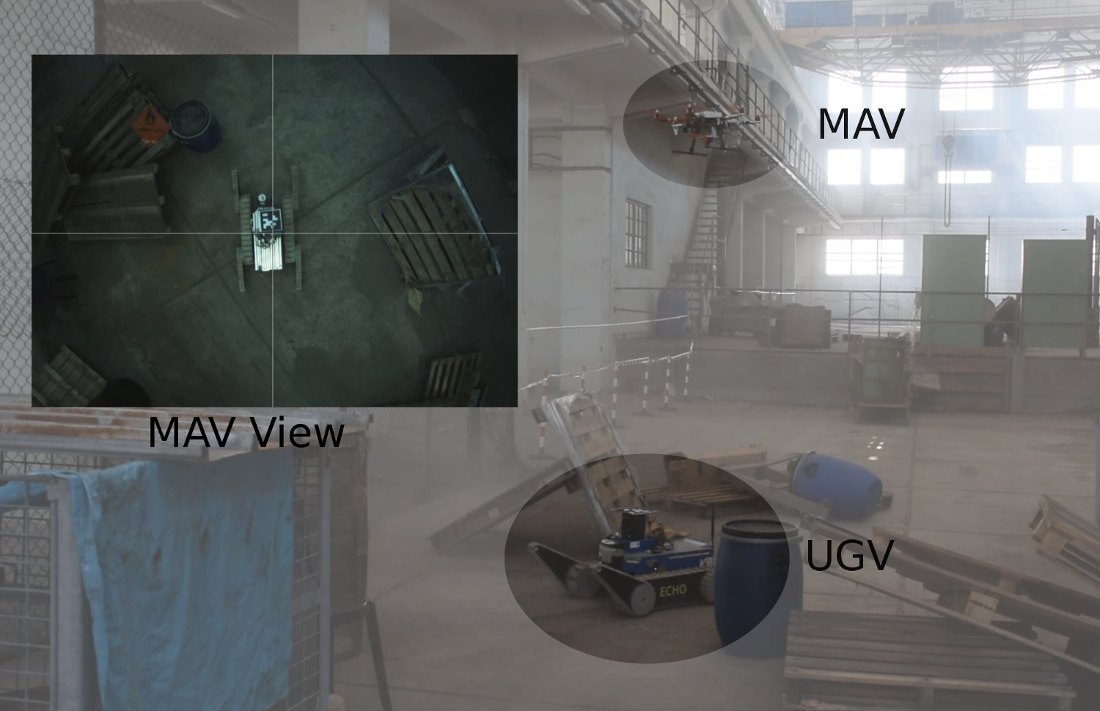}}
\caption{Experimental site Mestre, Italy}\label{fig:mestre}
\end{subfigure}
\hfill
\begin{subfigure}[c]{0.3\textwidth}
{\includegraphics[height=5cm]{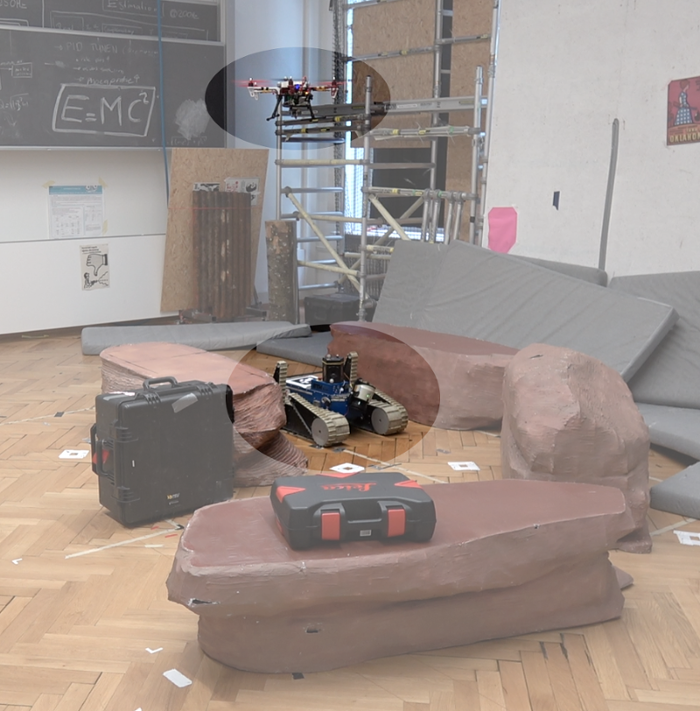}}
\caption{Experimental site Zurich, Switzerland}\label{fig:vicon_room}
\end{subfigure}
\hfill
\begin{minipage}[c]{0.2\textwidth}
\begin{subfigure}[t]{\linewidth}
\resizebox{\textwidth}{!}{\includegraphics{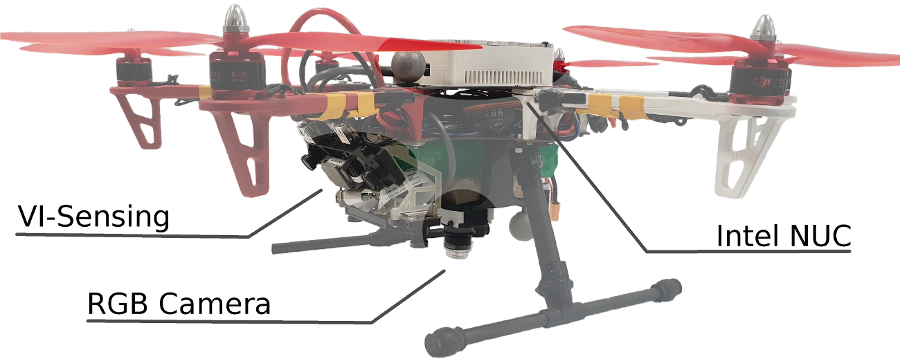}}
\caption{\ac{MAV}}\label{fig:MAV}
\end{subfigure}\\[\baselineskip]
\begin{subfigure}[b]{\linewidth}
\resizebox{\textwidth}{!}{\includegraphics{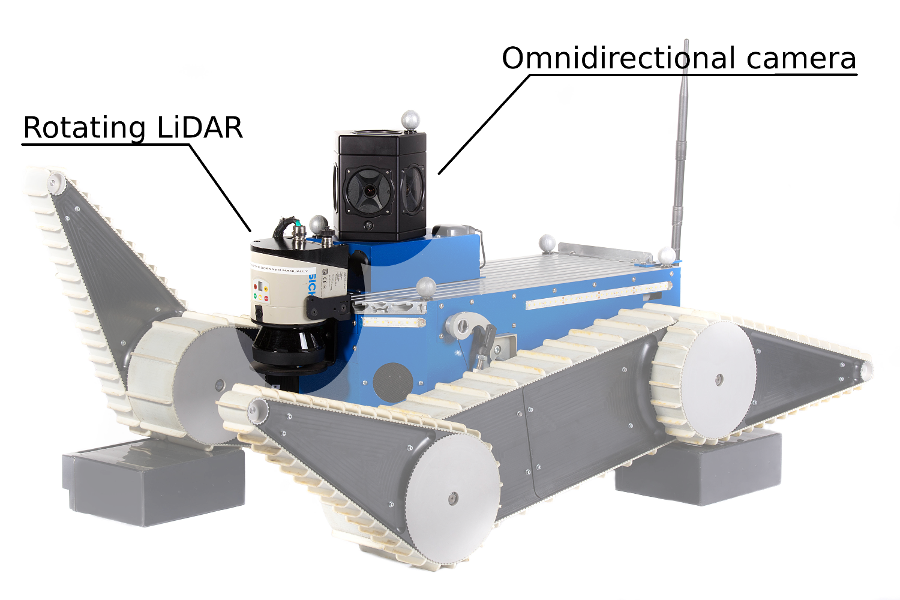}}
\caption{\ac{UGV}}\label{fig:nifti}
\end{subfigure}
\end{minipage}
\caption{Experimental set-up: \subref{fig:mestre} Experiment in disaster scenario in Italy, the \ac{MAV} autonomously servoes over a \ac{UGV} and provides third-person view for teleoperation. \subref{fig:vicon_room} experiment in motion capture room in Zurich. \subref{fig:MAV}~The custom designed \ac{MAV} we used is fitted with Visual-Inertial sensing for \ac{VIO}, a downward facing PointGrey Chameleon 3 camera @ $3.2MP$ for \ac{UGV} detection, and an Intel Core i7-7567U CPU @ $3.50GHz$. \subref{fig:nifti}~The used \ac{UGV} with Omnidirectional camera, and rotating \emph{2D} LiDAR to produce \emph{3D} scans.}
\vspace{-1.0em}
\label{fig:experiments}
\end{figure*}
The \ac{MAV} used in our experiments is a custom build hexacopter based on the DJI Flamewheel F550, and illustrated in Fig.~\ref{fig:MAV}.
It is equipped with a downward facing  Chameleon 3 camera with a resolution of 3.2 Megapixels @ $20Hz$ which is used both for the visual servoing and providing the third-person view\footnote{For the tag-based, and the feature-based detection, we use the implementations from \href{https://github.com/RIVeR-Lab/apriltags_ros}{https://github.com/RIVeR-Lab/apriltags\_ros} and \href{https://github.com/introlab/find-object}{https://github.com/introlab/find-object} respectively.}.
Furthermore, it is equipped with a variable set of one or two stereo-vision pairs integrated with an IMU for \ac{VIO}.
The stereo pairs are mounted 45$^{\circ}$ and 90$^{\circ}$ with respect to the horizontal.
All processing is done onboard the \ac{MAV} on an Intel Core i7-7567U CPU @ $3.50GHz$.
The flight time of the \ac{MAV} with an initially fully charged battery is approximately $15$ minutes.
The \ac{UGV} used for the LiDAR-based mapping is a tracked vehicle, equipped with encoders, IMU, and a sweeping LiDAR producing full \emph{3D} scans @~$1/3Hz$, as illustrated in Fig.~\ref{fig:nifti}.
The first-person view video stream is produced by a Ladybug 360 Degree camera @ $5Hz$, allowing the user to have an omnidirectional view around the robot.
The maximum speed of the \ac{UGV} is $0.6 m/s$, and is commonly operated at $0.3 m/s$.
All processing is done onboard using an Intel i7-4770T @ $2.5GHz$.

The Operator Control station is interfacing with the robots via WiFi using ROS and displays the video streams from the \ac{UGV}.
It also gives access to the \ac{MAV} third-person view requesting function, and displaying of its video stream.

Firstly, we perform structured experiments in a mock-up environment with a ground-truth pose tracking system for all robots, see Fig.~\ref{fig:vicon_room}.
We evaluate the effectiveness of the visual detection system under varying perceptual conditions, i.e., different payloads on robot, than on detection template, and inside / outside operation using the same detection template.
Also, we verify the minimum altitude of the \ac{MAV} for stable visual servoing.

Then, we map the room with our \ac{UGV}, request the \ac{MAV} between different locations, based on the LiDAR-based map, and evaluate the performance.
Since, we presently have only one \ac{UGV} available, we simulate the multi-\ac{UGV} scenario, by building a first map with the \ac{UGV} from the starting point, and then placing a visual target at its location.
Then, we use the \ac{UGV} to drive another way from the starting location and build a second map.
The locations of the \ac{UGV}s were chosen to maximize the distance between them in the experiment room, i.e., $5m$.
Both maps are registered based on their initial alignment.
We then let the \ac{MAV} travel between the visual target that simulates the first \ac{UGV} and the \ac{UGV}'s location of the second run.
In this experiment, we evaluate the request success, and the drift between the two pose estimations, i.e., \ac{VIO} for the \ac{MAV}, and LiDAR-based SLAM for the \ac{UGV}.

Secondly, we test the full system in both the mock-up scenario, and a realistic disaster scenario within the TRADR project review in Mestre, Italy, see Fig.~\ref{fig:mestre}.
The site consisted of a large decommissioned working hall, featuring various industrial installations, and obstacles on the ground.
Here, we demonstrate the full functionality of requesting the \ac{MAV}, transferring to the \ac{UGV} and supporting teleoperation with a third-person view by automatic visual servoing.
\subsection{Results}
\begin{figure}
\centering
\begin{tabular}[c]{cccc}
\begin{subfigure}[b]{0.23\columnwidth}
{\includegraphics[width=\textwidth]{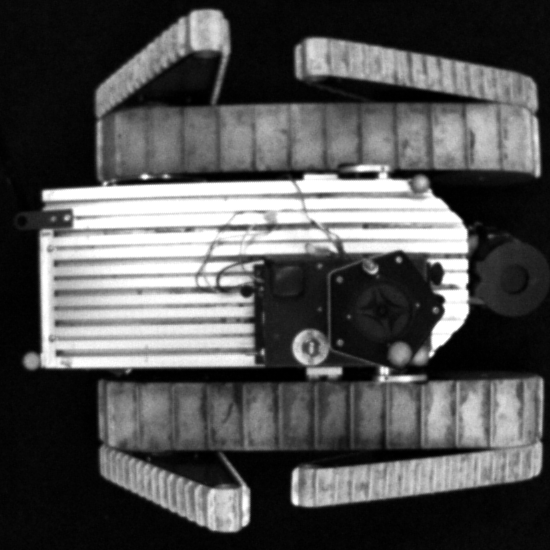}}
\label{fig:template_a}
\end{subfigure}&
\begin{subfigure}[b]{0.23\columnwidth}
{\includegraphics[width=\textwidth]{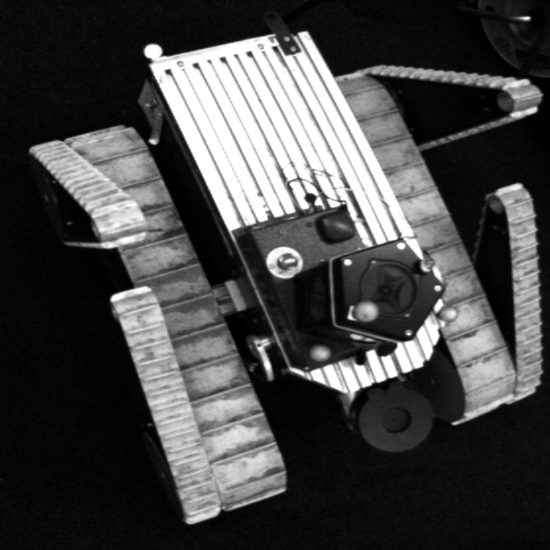}}
\label{fig:template_b}
\end{subfigure}&
\begin{subfigure}[b]{0.23\columnwidth}
{\includegraphics[width=\textwidth]{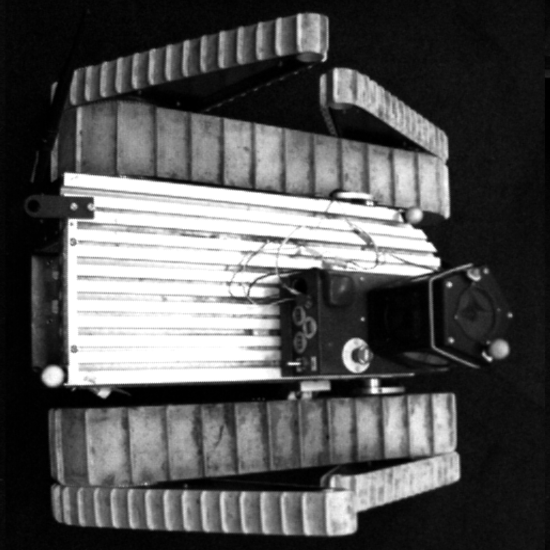}}
\label{fig:template_c}
\end{subfigure}&
\begin{subfigure}[b]{0.23\columnwidth}
{\includegraphics[width=\textwidth]{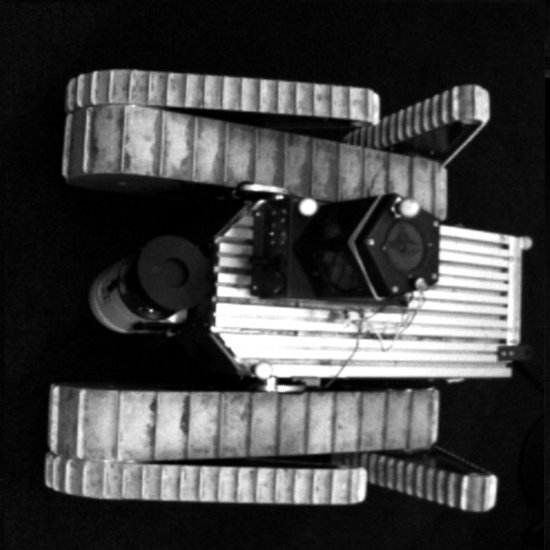}}
\label{fig:template_d}
\end{subfigure}\\
\begin{subfigure}[b]{0.23\columnwidth}
{\includegraphics[width=\textwidth]{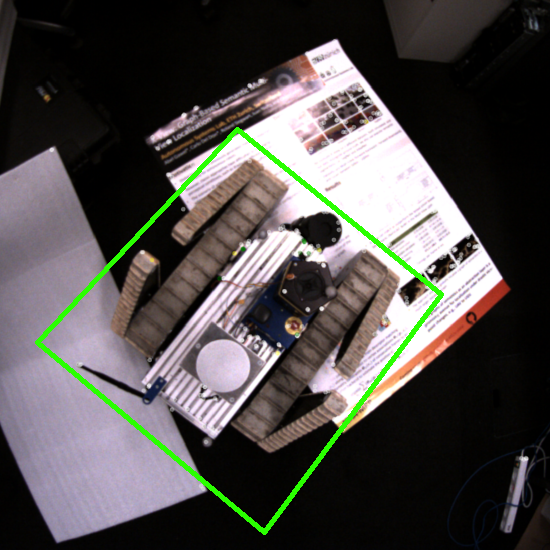}}
\label{fig:detected_a}
\end{subfigure}&
\begin{subfigure}[b]{0.23\columnwidth}
{\includegraphics[width=\textwidth]{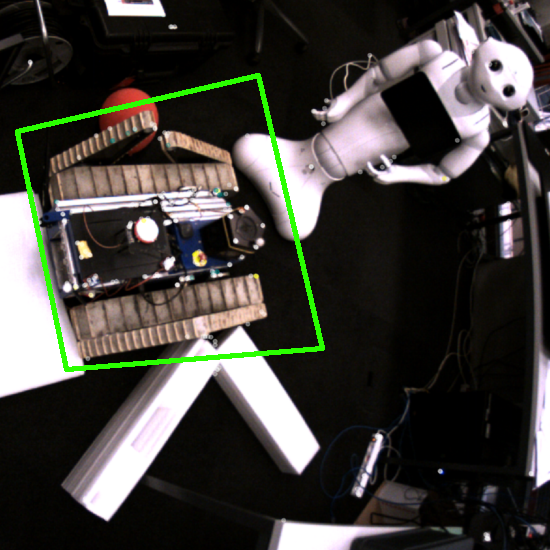}}
\label{fig:detected_b}
\end{subfigure}&
\begin{subfigure}[b]{0.23\columnwidth}
{\includegraphics[width=\textwidth]{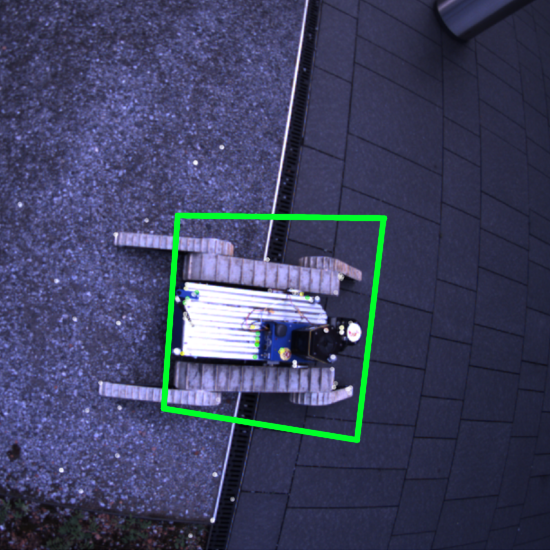}}
\label{fig:detected_c}
\end{subfigure}&
\begin{subfigure}[b]{0.23\columnwidth}
{\includegraphics[width=\textwidth]{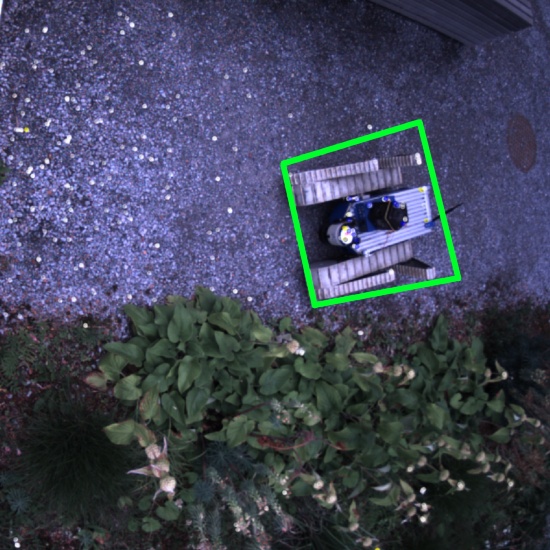}}
\label{fig:detected_d}
\end{subfigure}\\
\vspace{-1.0em}
\end{tabular}  
\centering
\caption{Sample images from the visual feature based detection: (top) Images of robot in database from different view-points. (bottom) Visual detection of robot equipped with different payloads and in different scene contexts. The green boxes indicate the estimated robot location.  }
\label{fig:mosaic}
\vspace{-1.0em}
\end{figure}

Given the hardware that we used in our experiments, an altitude of $2m$ above the servoing target showed to be sufficient to follow the \ac{UGV} in the proposed servoing mode.
However, visual detection both using visual features and Apriltags also performed well at greater heights up to the greatest tested height of $5m$.
While the Apriltag detection is very robust given visibility of the tag, also the visual feature based detection shows to work reliably.
Both detection regimes run onboard in real-time.
Given that the \ac{MAV} stays within the height range and maximum radius $r_{max}$, both receive at least one correct detection per second without false detections.
Samples of the visual feature based detection are illustrated in Fig.~\ref{fig:mosaic}.
However, using feature-based detection, the pose of the robot cannot always be precisely estimated.
Variations in the detected bounding box lead to these deviations, as also illustrated in Fig.~\ref{fig:mosaic}~(bottom row: middle left, and middle right).
\setlength{\tabcolsep}{1pt}
\renewcommand{\arraystretch}{0.01}

\begin{figure}
\centering
\includegraphics[width = \columnwidth]{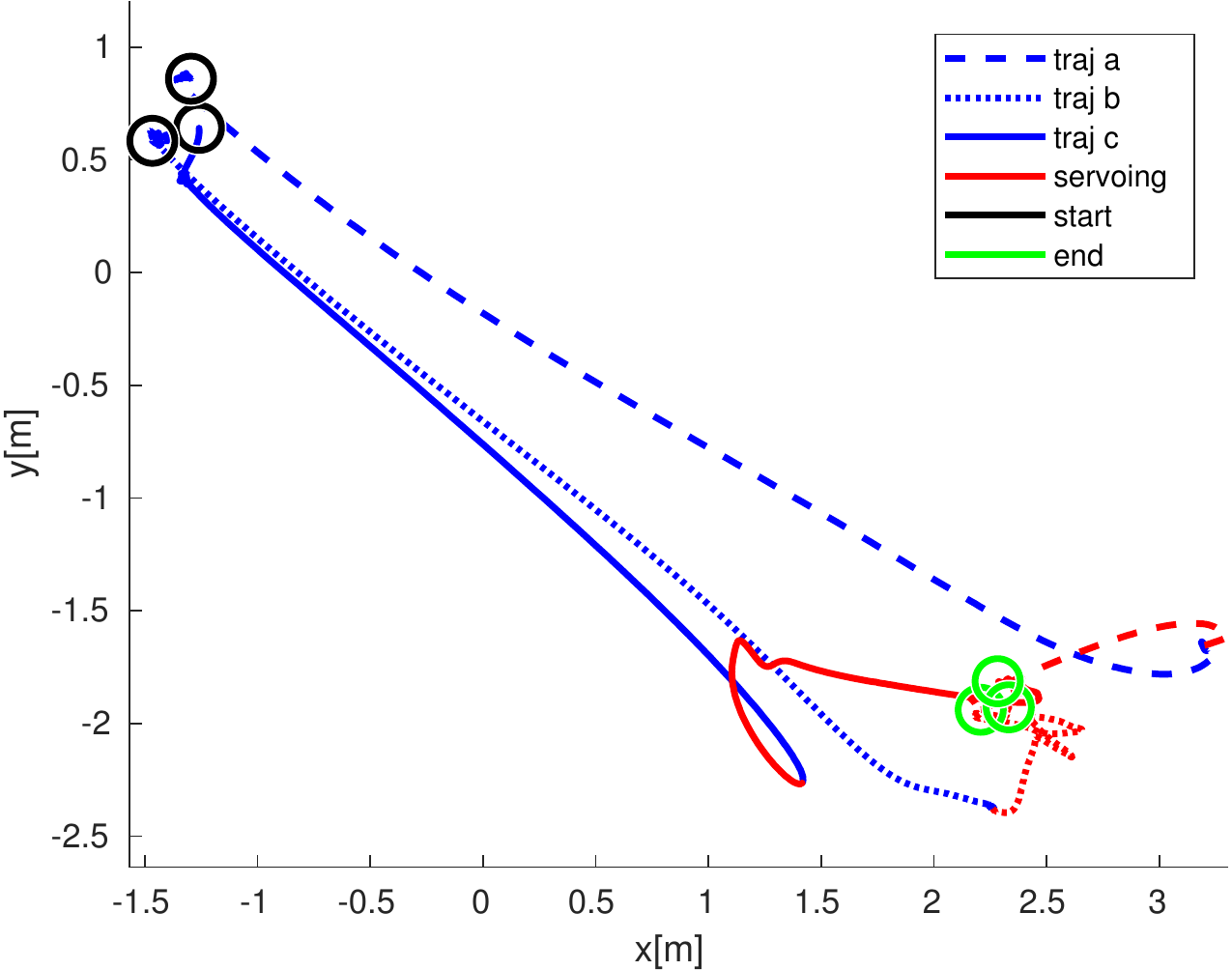}
\caption{Exemplary trajectories of the \ac{MAV} recorded from the motion capturing system. The \ac{MAV} is started from hovering at the black circled locations and send to various locations around the goal position (green circles). The visual servoing is able to compensate for displacements of up to $1.2m$ in our current set-up. Here, the trajectories traversed using \ac{VIO} are highlighted in blue, and the final visual servoing corrections in red.}
\vspace{-1.5em}
\label{fig:plot}
\end{figure}

The experiment on repeated transfers between two targets showed satisfying performance.
In series of $10$ transfers between the targets, the \ac{MAV} was always able to detect the new target and return to visual servoing above it.
We measured a $2cm$ displacement of the \ac{UGV} locations with respect to the ground truth locations, and an average displacement in $x,y$-coordinates between the \ac{MAV}'s position estimate and the ground-truth of $17.6cm$, with maximum displacements of up to $28.6cm$.
Furthermore, we evaluate the acceptable displacement from the target location.
In our experimental set-up, the \ac{MAV} is able to compensate for displacements of $1.2m$, i.e., four times the maximal error, from the target location through visual servoing, as illustrated in Fig.~\ref{fig:plot}.
Despite equal goal positions, the flown trajectories differ due to varying initializations of the \ac{VIO} in the starting location, and drift behaviour during the maneuver execution.

\begin{figure}[!htb]
\centering
\begin{subfigure}[c]{1.0\columnwidth}
{\includegraphics[width=0.95\columnwidth]{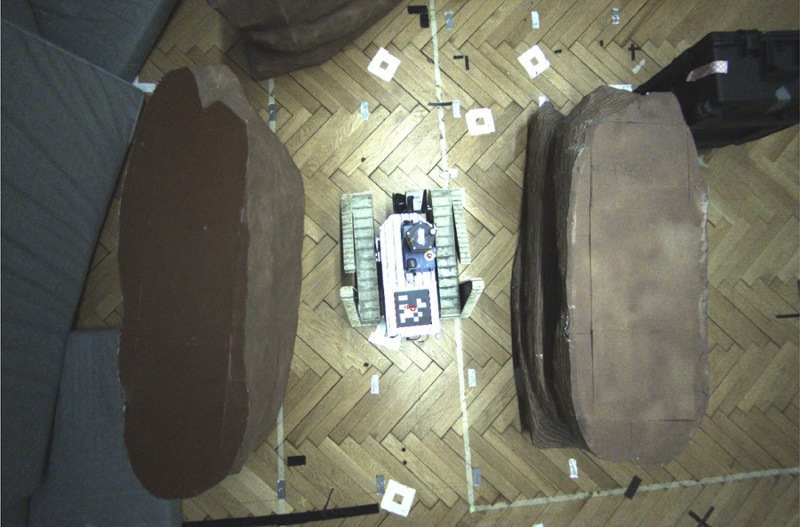}}
\caption{\ac{MAV} view}\label{fig:third-person}
\end{subfigure}
\begin{subfigure}[c]{1.0\columnwidth}
{\includegraphics[width=0.95\columnwidth]{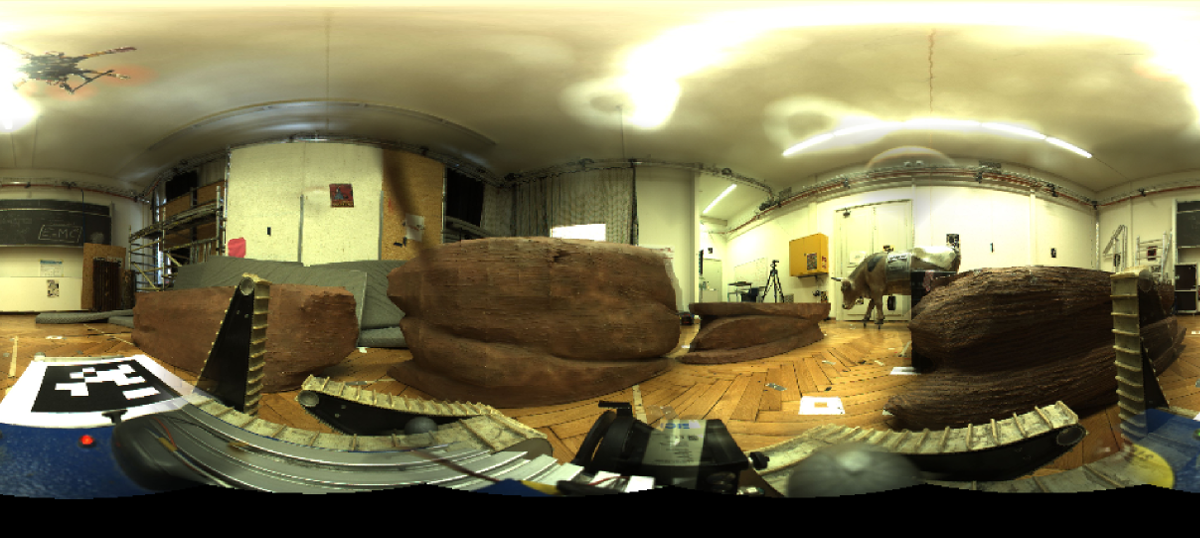}}
\caption{\ac{UGV} view}\label{fig:first-person}
\end{subfigure}
\caption{Third-person view from \ac{MAV}~(\subref{fig:third-person}) and panoramic stitched first-person view from \ac{UGV}~(\subref{fig:first-person}).}
\vspace{-1.5em}
\label{fig:views}
\end{figure}

Also, we successfully perform the requesting of the \ac{MAV} to a \ac{UGV}, and guiding it through difficult passages by providing a third-person view.
Exemplary views from \ac{MAV} and \ac{UGV} perspective are illustrated in Fig.~\ref{fig:views}.

Finally, the system shows to be easily controllable as no trained pilot is required to control the \ac{MAV}.
The system was controlled by several \ac{UGV} pilots that were not trained in \ac{MAV} piloting, and successfully completed the experiment parcours with the \ac{MAV} autonomously following the \ac{UGV}.

\subsection{Discussion}
The experimental results show that an interfacing between the localization systems of \ac{MAV}s with \ac{UGV}s, and integration with local systems, such as visual servoing can build an efficient real-time collaborative team of robots in GPS-denied environments. 
Several practical concerns have to be taken into account in designing such systems.
Our choice for hovering the \ac{MAV} in a fixed location within a large margin $r_{max}=0.2m$ above the \ac{UGV} is partly due to using a fixed downward facing camera for both visual servoing and providing the third-person view.
Having a constant following of the \ac{UGV} results in a unsteady image, rendering teleoperation cumbersome.
Low tolerances on the location above the \ac{UGV} can furthermore be difficult to accomplish by the \ac{MAV} controller.
Here, a camera equipped with a gimbal could give a steady image of the target for teleoperation, and enable lower margins in the hovering location, enabling a more steady image stream of the \ac{UGV}.
Furthermore, tight margins cannot be achieved when using visual feature-based detection, causing jumps in the detection.

While the drift in the LiDAR-based localization of the \ac{UGV}s is negligible in the evaluated scales, the \ac{VIO} is showing larger drifts.
This is partly due to tuning our flight controller for aggressive flying maneuvers, leading to quick transfer times between targets.
While our experiments are successful throughout for returning to visual servoing above the new target, less aggressive maneuvers could decrease the drift further.
However, our present system can handle up to four times the maximum drift measured in the experiments.

Furthermore, we demonstrated our \ac{MAV} system for transferring between multiple \ac{UGV}s, but it is not limited to this.
The visual feature based detection enables our system to be allocated further targets within the map during the mission, e.g., locations of interest for periodic inspection.
Therefore, the task of the \ac{MAV} can be extended to periodic inspection using the same building blocks.
Another option, however not covered by this work, is the possibility to add active localization sensors such as RFID tags if the visual detection is challenged by occlusions or other perceptually difficult situations.
%
%
%

%
%
%
%
%
%
%
\section{Conclusion}
\label{sec:conclusion}
In this paper we presented a system for effective collaborative sensing for \ac{MAV}s and \ac{UGV}s based on combining global and local localization in GPS-denied \ac{SaR} environments. 
We show that we can effectively use an \ac{MAV} as flying camera to support multiple \ac{UGV} operators in teleoperation, by providing third-person views.
The proposed functionality of global transferring between multiple targets shows to work reliably throughout our experiments.
Our integration and evaluation reveals the constraints on each module and compromises to obtain a working reliable system.
For instance, the hovering strategy to favor better detection and image quality for the teleoperation task given the use of the same camera for both objectives. 
   
In future work, it would be interesting to add collision avoidance to the \ac{MAV} to lift the assumption of free space above the \ac{UGV}.
Furthermore, exchanging the rigid attachment of the downward facing camera with a gimbal could be a beneficial addition to the system.

\section{Acknowledgement}
This work was supported by European Union's Seventh Framework Program for research, technological development and demonstration under the TRADR project No. FP7-ICT-609763, and by the National Center of Competence in Research (NCCR) Robotics through the Swiss National Science Foundation.
The authors would like to thank Rik B\"ahnemann, Marius Grimm, Alexander Millane, Helen Oleynikova, Dr. Zachary Taylor, and Renaud Dub\'e for their valuable collaboration and support.

\bibliographystyle{IEEEtranN}

\bibliography{eth}

\begin{acronym}
\acro{MAV}{Micro Aerial Vehicle}
\acro{UAV}{Unmanned Aerial Vehicle}
\acro{UGV}{Unmanned Ground Vehicle}
\acro{TRADR}{``Long-Term Human-Robot Teaming for Robots Assisted Disaster Response''}
\acro{SaR}{Search and Rescue}
\acro{MPC}{Model Predictive Controller}
\acro{VIO}{Visual Inertial Odometry}
\end{acronym}

\end{document}

%% file: preface.tex

\thispagestyle{empty}

{\setlength{\parindent}{0cm}
\textcopyright 2018 IEEE. Personal use of this material is permitted. Permission from IEEE must be obtained
for all other uses, in any current or future media, including reprinting/republishing this material
for advertising or promotional purposes, creating new collective works, for resale or redistribution
to servers or lists, or reuse of any copyrighted component of this work in other works.
}
\\

{\setlength{\parindent}{0cm}
Pre-print of article that will appear in the 2018 IEEE International Symposium on Safety, Security and Rescue Robotics (SSRR).
}
\\

{\setlength{\parindent}{0cm}
Please cite this paper as:
}
\\

{\setlength{\parindent}{0cm}
A. Gawel, Y. Lin, T. Koutros, R. Siegwart, and C. Cadena. (2018).\\ "Aerial-Ground collaborative sensing: Third-Person view for teleoperation" in IEEE International Symposium on Safety, Security and Rescue Robotics (SSRR), 2018.
}
\\

{\setlength{\parindent}{0cm}
bibtex:
}
\\

{\setlength{\parindent}{0cm}
@inproceedings\{gawel2018aerial,\\
  \phantom{x} title=\{Aerial-Ground collaborative sensing: Third-Person view for teleoperation\},\\ 
  \phantom{x} author=\{Gawel, Abel and Lin, Yukai and Koutros, Theodore and Siegwart, Roland and Cadena, Cesar\},\\ 
  \phantom{x} booktitle={Safety, Security and Rescue Robotics (SSRR), IEEE International Symposium on\},\\
  \phantom{x} year=\{2018\}
  \\ 
\}
}
